\documentclass[runningheads]{llncs}
\usepackage[T1]{fontenc}
\usepackage{graphicx,verbatim}
\usepackage{amsmath}
\usepackage{amssymb}
\usepackage{booktabs}
\usepackage{multirow}
\usepackage{pifont}
\usepackage{dsfont}
\usepackage{arydshln}
\usepackage{array}
\usepackage{tabularx}
\usepackage{marvosym}
\begin{document}
\title{Text as Illumination: Spatial Contrastive Retinex Learning for Language-guided Medical Image Segmentation}
%
\author{Jian Shi\inst{1}\index{Shi, Jian},
Cheng Zhen\inst{1}\index{Zhen, Cheng},
Pingping Zhang\inst{2}\textsuperscript{(\Letter)}\index{Zhang, Pingping},
Rui Xu\inst{1}\index{Xu, Rui},
Yanan Lv\inst{3}\index{Lv, Yanan},
Yili Ma\inst{3}\index{Ma, Yili},
Huan Bi\inst{4}\index{Bi, Huan},
Haojie Li\inst{5}\index{Li, Haojie},
and Huchuan Lu\inst{2}\index{Lu, Huchuan}}

\authorrunning{Jian Shi et al.}
%
\institute{School of Software Technology \& DUT-RU International School of Information Science and Engineering, Dalian University of Technology, Dalian, China
\and
School of Future Technology, Dalian University of Technology, Dalian, China
\and
Central Hospital of Dalian University of Technology, Dalian, China
\and
Cancer Hospital of Dalian University of Technology, Dalian, China
\and
College of Computer Science and Engineering, Shandong University of Science and Technology, Qingdao, China
\\
\email{zhpp@dlut.edu.cn}
}


\maketitle
\begin{abstract}
Language-guided Medical Image Segmentation (LMIS) has shown great potential to improve the delineation of anatomical structures and lesions by integrating clinical textual information.
Existing methods generally rely on either implicit interaction between textual and visual features or auxiliary coarse-grained supervision for cross-modal alignment.
However, these methods lack explicit and fine-grained constraints to ensure semantic consistency, causing a mismatch between language and the segmentation outputs.
To address this issue, we propose Text-as-Illumination Retinex Network (TIRNet), a novel Retinex-inspired framework that treats text embeddings as semantic illumination for feature modulation, thereby improving semantic
consistency in LMIS.
TIRNet introduces two key blocks integrated at each decoder stage: (1)~the Retinex-inspired Text Modulation Block (RTMB), which employs positive and negative illumination maps to enhance text-relevant foreground features and suppress background interference; and (2)~the Consistent Detail Compensation Block (CDCB), which selectively recovers high-frequency details via a consistency-gated mechanism conditioned on illumination reliability.
Furthermore, we propose a Multi-Scale Illumination Supervision Loss (MSIS-Loss), comprising a Region-Grounded Contrastive Loss (RGC-Loss) that enforces cross-modal similarity to be concentrated in text-relevant foreground regions and suppressed in background regions, and a Background Suppression Loss (BS-Loss) that provides pixel-level supervision for negative illumination maps, jointly ensuring a precise cross-modal alignment at each decoder stage.
Extensive experiments on the MosMedData+ and QaTa-COV19 datasets demonstrate that TIRNet achieves state-of-the-art performance in LMIS.
The code is available at: https://github.com/anaanaa/TIRNet.


\end{abstract}
%
%
\section{Introduction}

Medical image segmentation serves as a fundamental yet challenging task in computer-aided diagnosis, focusing on the precise delineation of key regions of interest, such as lesions and anatomical structures~\cite{bozorgpour2025cenet,zhu2024selfreg}.
Although purely vision-based deep learning methods have achieved considerable progress, their success relies on large-scale datasets with expert-level pixel-wise annotations.
However, acquiring these annotations is expensive and labor-intensive, leading to a scarcity of high-quality labeled data.
Notably, routine clinical reports generated in standard medical practice provide rich complementary semantic priors.
Motivated by this observation, Language-guided Medical Image Segmentation (LMIS)~\cite{huang2024cross,yang2022lavt,zhong2023ariadne} provides a promising way to leverage textual priors.

\begin{figure}
\includegraphics[width=0.98\textwidth]{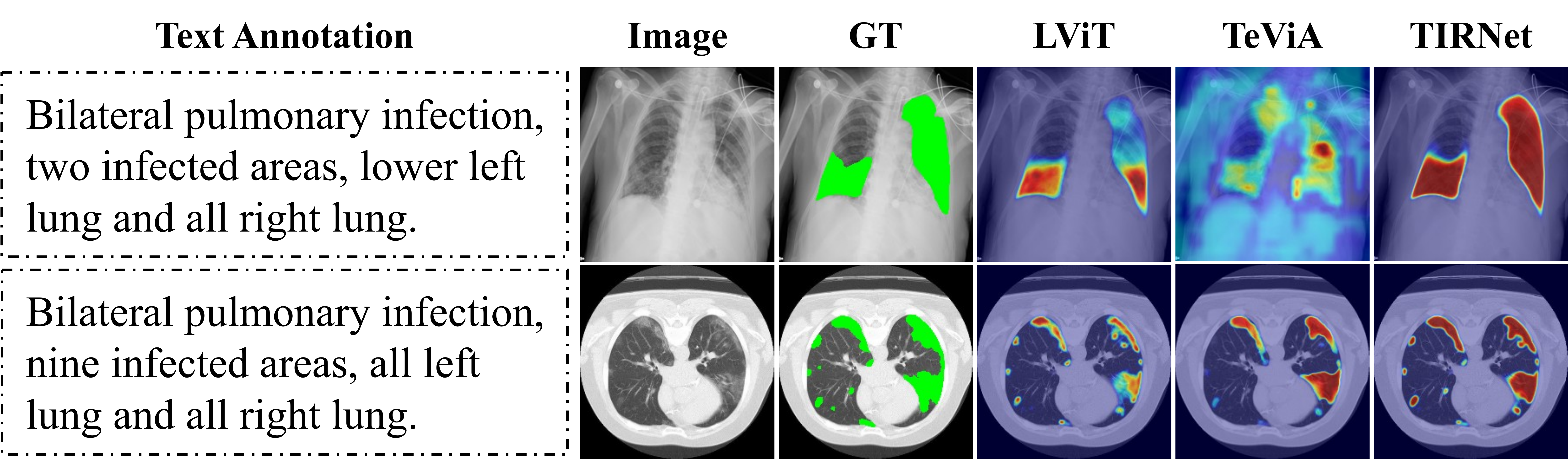}
\caption{Limitations of existing methods. Both implicit interaction (LViT~\cite{li2023lvit}) and coarse supervision (TeViA~\cite{11133474}) methods fail to effectively focus on foregrounds, while our TIRNet achieves accurate localization.} \label{intro}
\end{figure}

Existing LMIS methods generally follow two representative paradigms: implicit interaction methods that fuse textual and visual features via feature merging or attention ~\cite{li2023lvit,guo2024common}, and coarse supervision methods that introduce auxiliary constraints for cross-modal alignment~\cite{11133474,zhou2025hybrid}.
However, the former lacks explicit constraints to ensure semantic consistency across modalities, while the latter provides only coarse-grained supervision, limiting its ability to model fine spatial correspondence.
For instance, LViT~\cite{li2023lvit} integrates text embeddings via a pixel-level attention module, but the interaction between textual and visual features remains implicit and lacks explicit region-level alignment.
TeViA~\cite{11133474} introduces explicit supervision by aligning foreground visual representations extracted from the segmentation head with text through a momentum-updated prototype. However, this coarse supervision is insufficient for modeling fine spatial correspondence between textual semantics and pixel-wise predictions.
As shown in Fig.~\ref{intro}, both LViT and TeViA struggle to effectively focus on the foreground regions.

To address the above issues, we present Text-as-Illumination Retinex Network (TIRNet), a novel framework for LMIS.
Specifically, TIRNet integrates two key blocks at each decoder stage.
The Retinex-inspired Text Modulation Block (RTMB) derives positive and negative illumination maps from cross-modal similarities to explicitly amplify foregrounds and suppress backgrounds.
The Consistent Detail Compensation Block (CDCB) selectively recovers high-frequency details via an illumination-guided consistency gate.
Furthermore, we propose a Multi-Scale Illumination Supervision Loss (MSIS-Loss) to enforce explicit alignment by employing a Region-Grounded Contrastive Loss (RGC-Loss) to maximize foreground-background margins and a pixel-level Background Suppression Loss (BS-Loss) to explicitly suppress background responses.
Extensive experiments on the MosMedData+ and QaTa-COV19 datasets demonstrate that TIRNet achieves state-of-the-art performance in LMIS.

Our main contributions are summarized as follows:
\begin{itemize}
\item We propose TIRNet for LMIS, which draws inspiration from Retinex theory by treating text embeddings as semantic illumination, enabling explicit foreground amplification and background suppression in decoder features.
\item We design the RTMB to derive complementary illumination maps for feature modulation, and the CDCB to selectively recover high-frequency details via an illumination-guided consistency gate.
\item We introduce a MSIS-Loss, combining RGC-Loss and BS-Loss, to enforce cross-modal semantic alignment.
\item Experiments on the MosMedData+ and QaTa-COV19 datasets demonstrate that TIRNet achieves superior performance over existing methods.
\end{itemize}
%
\section{Method}

\begin{figure}
\includegraphics[width=0.98\textwidth]{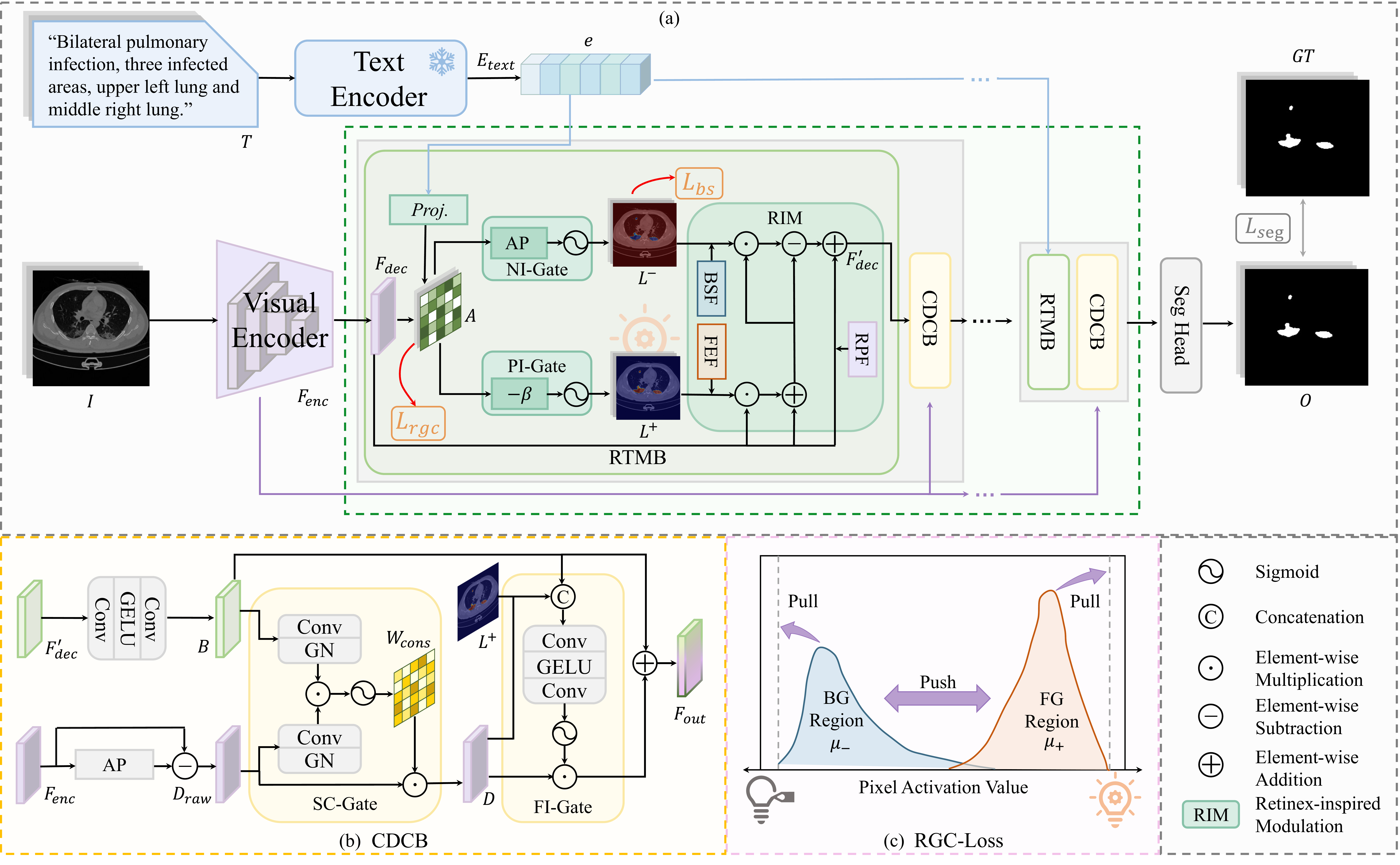}
\caption{Overview of the TIRNet. It integrates RTMB and CDCB into each decoder stage. The RGC-Loss maximizes cross-modal similarity in text-relevant foregrounds and suppresses background activations to enlarge the foreground-background margin.} \label{fig:overview}
\end{figure}

As shown in Fig.~\ref{fig:overview}, TIRNet adopts a U-shaped encoder-decoder architecture.
A CNN-based visual encoder extracts multi-scale visual features $\{\mathbf{F}_{\mathrm{enc}}^{s}\}_{s=1}^{4}$, while a frozen CLIP~\cite{radford2021learning} text encoder extracts token-level text features $\mathbf{E}_{\mathrm{text}} \in \mathbb{R}^{L \times D}$, which are refined by a 1D convolution and mean-pooled into a global semantic embedding $\mathbf{e} \in \mathbb{R}^{D}$.
At each decoder stage, the RTMB uses $\mathbf{e}$ as semantic illumination to modulate features, which are then fed to the CDCB to recover high-frequency details from the corresponding encoder stage.
During training, similarity maps from each RTMB are supervised by the proposed MSIS-Loss.

\subsection{Retinex-inspired Text Modulation Block}
\label{sec:rtcm}

Retinex theory~\cite{land1977retinex} models an image as $\mathbf{S} = \mathbf{R} \odot \mathbf{L}$, where the reflectance $\mathbf{R}$ represents the intrinsic structural content and the illumination $\mathbf{L}$ modulates spatial visibility.
Recent Retinex-inspired methods have shown that Retinex-based decomposition and illumination-aware modeling can effectively guide feature enhancement, ranging from low-level image restoration to multispectral object detection~\cite{cai2023retinexformer,LiuGCWYSM25}.
Inspired by this illumination-guided representation mechanism, we propose the Retinex-inspired Text Modulation Block (RTMB), which adapts Retinex decomposition to cross-modal feature modulation. In RTMB, the visual decoder feature $\mathbf{F}_{\mathrm{dec}}$ acts as the reflectance $\mathbf{R}$, while the text embedding serves as a semantic illumination field $\mathbf{L}$.
This text-conditioned modulation enhances target regions that are semantically aligned with the text embedding while suppressing irrelevant background responses, producing the modulated decoder feature $\mathbf{F}_{\mathrm{dec}}^{\prime}$ with an improved foreground-background separability.

At the $s$-th decoder stage, the RTMB takes the decoder feature
$\mathbf{F}_{\mathrm{dec}}^{(s)} \in \mathbb{R}^{C_s \times H_s \times W_s}$
and the global text embedding $\mathbf{e}$ as inputs.
For notational simplicity, we omit the superscript $(s)$ in the following equations.
Given the decoder feature  $\mathbf{F}_{\mathrm{dec}} \in \mathbb{R}^{C \times H \times W}$ and $\mathbf{e}$, we firstly project $\mathbf{e}$ into the image feature space via a learnable linear projection  $\mathbf{W}_{p} \in \mathbb{R}^{C \times D}$:

\begin{equation}
    \mathbf{t} = \mathbf{W}_{p}\mathbf{e} ,
\quad \mathbf{t} \in \mathbb{R}^{C}.
\end{equation}

Then we compute the text-guided relevance score at each spatial location:
\begin{equation}
A_{h,w} =
\left\langle
\frac{\mathbf{f}_{h,w}}{\|\mathbf{f}_{h,w}\|_2},
\frac{\mathbf{t}}{\|\mathbf{t}\|_2}
\right\rangle,
\label{eq:sim_map}
\end{equation}
where $\mathbf{f}_{h,w} \in \mathbb{R}^{C}$ is the decoder feature vector at spatial location $(h,w)$, and $\langle \cdot,\cdot \rangle$ is the inner product.
The map $\mathbf{A} \in \mathbb{R}^{1 \times H \times W}$ aggregates channel-wise normalized visual-text correlations into a semantic relevance score at each spatial location.

Based on this semantic relevance map $\mathbf{A}$, we derive two complementary illumination maps through two parallel gating pathways.
The Positive Illumination Gate (PI-Gate) applies a
$3{\times}3$ Average-Pooling
$\mathrm{(AP)}$ followed by a sigmoid activation function $\sigma$ to produce a spatially coherent foreground illumination:
\begin{equation}
    \mathbf{L}^{+} = \sigma\!\bigl(\mathrm{AP}(\mathbf{A})\bigr)
    \in [0,1]^{1 \times H \times W}.
    \label{eq:pos_illum}
\end{equation}

Conversely, the Negative Illumination Gate (NI-Gate) inverts the similarity with a learnable sharpness $\beta$ to capture text-irrelevant backgrounds:
\begin{equation}
    \mathbf{L}^{-} = \sigma\!\bigl(-\beta \mathbf{A}\bigr)
    \in [0,1]^{1 \times H \times W},
    \label{eq:neg_illum}
\end{equation}
where $\beta$ is strictly positive.
The negation ensures that low-relevance
locations receive high activations in $\mathbf{L}^{-}$, while
the learnable $\beta$ adaptively controls the
foreground--background transition sharpness.

\noindent\textbf{Retinex-inspired Modulation.}
The decoder feature is modulated as:
\begin{equation}
    \mathbf{F}_{\mathrm{dec}}^{\prime} =
    \mathbf{F}_{\mathrm{dec}} \odot
    \bigl(\mathbf{1} + \rho \mathbf{L}^{+}\bigr) \odot
    \bigl(\mathbf{1} - \kappa \mathbf{L}^{-}\bigr) +
    \gamma \mathbf{F}_{\mathrm{dec}},
    \label{eq:rtcm_mod}
\end{equation}
where $\rho$, $\kappa$, and $\gamma$ denote the learnable Foreground Enhancement Factor (FEF), Background Suppression Factor (BSF), and Residual Preservation Factor (RPF).

\subsection{Consistent Detail Compensation Block}
\label{sec:cdc}

Although the RTMB effectively disentangles the foreground and background, the illuminated regions still lack high-frequency details essential for precise boundary delineation due to feature downsampling.
To address this issue, the Consistent Detail Compensation Block (CDCB) selectively recovers fine-grained structural information from the encoder under the guidance of illumination reliability.

Given the refined feature
$\mathbf{F}_{\mathrm{dec}}^{\prime}$, we derive a base representation
$\mathbf{B} = f_{\mathrm{base}}(\mathbf{F}_{\mathrm{dec}}^{\prime})$ via a $3{\times}3$ conv--GELU--$3{\times}3$ conv sequence.
To isolate high-frequency details from the encoder
feature $\mathbf{F}_{\mathrm{enc}}$, we compute the residual $\mathbf{D}_{\mathrm{raw}}$ between the original feature and its low-pass filtered counterpart, which extracts edge and texture information complementary to the decoder stream~\cite{yan2023transy}:
\begin{equation}
    \mathbf{D}_{\mathrm{raw}} = \mathbf{F}_{\mathrm{enc}}
    - \mathrm{AP}(\mathbf{F}_{\mathrm{enc}}).
    \label{eq:detail}
\end{equation}

Then, we introduce a Semantic Consistency Gate
(SC-Gate) that selectively integrates encoder details semantically aligned with the decoder representation:
\begin{equation}
    \mathbf{W}_{\mathrm{cons}} = \sigma\!\Bigl(
    \delta \sum_{c=1}^{C}
    \bar{\mathbf{q}}_c \odot \bar{\mathbf{k}}_c\Bigr),
    \label{eq:consistency}
\end{equation}
where $\bar{\mathbf{q}} = \mathrm{GN}(g_q(\mathbf{B}))$ and
$\bar{\mathbf{k}} = \mathrm{GN}(g_k(\mathbf{D}_{\mathrm{raw}}))$.
$g_q$ and $g_k$ are $1 \times 1$ convolution layers. $\mathrm{GN}(\cdot)$ denotes Group Normalization~\cite{wu2018group}.
$\delta$ is a learnable parameter.
Consequently, the modulated detail is obtained as $\mathbf{D} = \mathbf{W}_{\mathrm{cons}} \odot \mathbf{D}_{\mathrm{raw}}$.

Finally, a Fusion Illumination Gate (FI-Gate) controls detail injection
using $\mathbf{L}^{+}$ as a cross-modal reliable signal:
\begin{equation}
    \mathbf{G} = \sigma\!\bigl(f_{\mathrm{gate}}
    ([\mathbf{B};\, \mathbf{D};\, \mathbf{L}^{+}])\bigr),
    \label{eq:gate}
\end{equation}
where $[\cdot;\cdot]$ is a channel-wise concatenation.
$f_{\mathrm{gate}}$ is a $3{\times}3$ conv--GELU--$1{\times}1$ conv sequence.
Guided by $\mathbf{L}^{+}$, the gate selectively preserves fine structural details within the target foreground while suppressing task-irrelevant background interference. The output aggregates the base representation with the gated detail as
$\mathbf{F}_{\mathrm{out}} = \mathbf{B} + \mathbf{G} \odot \mathbf{D}$.

\subsection{Multi-Scale Illumination Supervision Loss}
\label{sec:loss}

To enforce cross-modal alignment across all decoder stages, we propose a Multi-Scale Illumination Supervision Loss (MSIS-Loss) over the stage-wise illumination cues $\{\mathbf{A}^{(s)}, \mathbf{L}^{-(s)}\}_{s=1}^{S}$.
The MSIS-Loss consists of two complementary terms: a Region-Grounded Contrastive Loss (RGC-Loss) and a Background Suppression Loss (BS-Loss).
Unlike conventional instance-level~\cite{chen2020simple,rui2025multi} or pixel-level~\cite{shi2022semantic,wang2021exploring} contrastive losses, the RGC-Loss aggregates cross-modal similarities within anatomically meaningful foreground and background regions.
This spatially structured supervision effectively bridges the granularity gap between global text semantics and local pixel predictions.
Specifically, we downsample the ground-truth to
$\mathbf{G}^{(s)}$ at scale $s$, partitioning the spatial
domain into a positive region
$\Omega_{+}^{(s)} = \{(i,j) \mid \mathbf{G}^{(s)}_{i,j} = 1\}$
and a negative region
$\Omega_{-}^{(s)} = \{(i,j) \mid \mathbf{G}^{(s)}_{i,j} = 0\}$.
We then compute the exponentiated similarity
$\mathbf{P}^{(s)} = \exp(\mathbf{A}^{(s)})$ and aggregate
the mean activation intensity within each region:
\begin{equation}
    \mu_{+}^{(s)} =
    \frac{\sum_{i,j} \mathbf{G}^{(s)}_{i,j}\,
    \mathbf{P}^{(s)}_{i,j}}{|\Omega_{+}^{(s)}|},
    \quad
    \mu_{-}^{(s)} =
    \frac{\sum_{i,j} (1{-}\mathbf{G}^{(s)}_{i,j})\,
    \mathbf{P}^{(s)}_{i,j}}{|\Omega_{-}^{(s)}|}.
    \label{eq:fg_bg}
\end{equation}

The contrastive loss maximizes foreground-background margins:
\begin{equation}
    \mathcal{L}_{\mathrm{rgc}}^{(s)} = -\log
    \frac{\mu_{+}^{(s)}}
    {\mu_{+}^{(s)} + \mu_{-}^{(s)}}.
    \label{eq:loss_cl}
\end{equation}

Furthermore, we formulate a BS-Loss that explicitly supervises $\mathbf{L}^{-(s)}$ with the inverse ground-truth:
\begin{equation}
    \mathcal{L}_{\mathrm{bs}}^{(s)} =
    \mathrm{BCE}\!\bigl(\mathbf{L}^{-(s)},\;
    1 - \mathbf{G}^{(s)}\bigr).
    \label{eq:loss_bg}
\end{equation}

This loss ensures that the negative illumination correctly captures the non-target background.

We aggregate the stage-wise losses as:
\begin{equation}
    \mathcal{L}_{\mathrm{rgc}} =
    \frac{1}{S}\sum_{s=1}^{S}\mathcal{L}_{\mathrm{rgc}}^{(s)},
    \quad
    \mathcal{L}_{\mathrm{bs}} =
    \frac{1}{S}\sum_{s=1}^{S}\mathcal{L}_{\mathrm{bs}}^{(s)}.
\end{equation}

The overall training loss is formulated as:
\begin{equation}
    \mathcal{L}_{\mathrm{total}} =
    \mathcal{L}_{\mathrm{seg}} +
    \lambda_{\mathrm{rgc}} \mathcal{L}_{\mathrm{rgc}} +
    \lambda_{\mathrm{bs}} \mathcal{L}_{\mathrm{bs}},
    \label{eq:total}
\end{equation}
where $\mathcal{L}_{\mathrm{seg}}$ is the
segmentation loss, combining a Dice loss and a binary cross-entropy
loss.
The loss weights are experimentally set to
$\lambda_{\mathrm{rgc}}{=}1.0$ and
$\lambda_{\mathrm{bs}}{=}0.5$.

\begin{table}[htbp]
\caption{Quantitative comparison on MosMedData+ and QaTa-COV19 datasets. The dashed line separates uni-modal methods (top) from multi-modal methods (bottom).}
\label{tab:comparison}
\centering
\renewcommand{\arraystretch}{1.1}
\resizebox{\textwidth}{!}{%
\begin{tabular}{@{}l*{10}{c}@{}}
\toprule
\multirow{2}{*}{\textbf{Method}} & \multirow{2}{*}{\textbf{Params (M)}} & \multirow{2}{*}{\textbf{Flops (G)}} & \multicolumn{4}{c}{\textbf{MosMedData+}} & \multicolumn{4}{c}{\textbf{QaTa-COV19}} \\
\cmidrule(lr){4-7}\cmidrule(lr){8-11}
 & & & $\mathrm{m\text{-}Dice}$ & $\mathrm{m\text{-}IoU}$ & $\mathrm{g\text{-}Dice}$ & $\mathrm{g\text{-}IoU}$ & $\mathrm{m\text{-}Dice}$ & $\mathrm{m\text{-}IoU}$ & $\mathrm{g\text{-}Dice}$ & $\mathrm{g\text{-}IoU}$ \\
\midrule
U-Net~\cite{ronneberger2015u} & 14.80 & 50.3 & 62.52 & 49.24 & 72.38 & 56.71 & 78.46 & 68.59 & 86.66 & 76.47 \\
U-Net++~\cite{zhou2018unet++} & 74.50 & 94.6 & 69.74 & 57.39 & 80.19 & 66.93 & 79.55 & 70.21 & 87.76 & 78.19 \\
nnUNet~\cite{isensee2021nnu} & 19.10 & 412.7 & 73.18 & 61.02 & 80.46 & 67.31 & 79.13 & 69.69 & 87.37 & 77.57 \\
Swin-UNet~\cite{cao2022swin} & 82.30 & 67.3 & 65.94 & 52.35 & 77.85 & 63.73 & 77.29 & 67.55 & 86.44 & 76.12 \\
\noalign{\vskip 2pt}\cdashline{1-11}[1.5pt/2pt]\noalign{\vskip 2pt}
GLoRIA~\cite{huang2021gloria} & 45.60 & 60.8 & 70.11 & 56.75 & 79.31 & 65.71 & 79.51 & 70.36 & 88.25 & 78.97 \\
LAVT~\cite{yang2022lavt} & 118.60 & 83.8 & 73.66 & 60.74 & 80.66 & 67.58 & 78.87 & 69.35 & 87.56 & 77.88 \\
TGANet~\cite{tomar2022tganet} & 19.80 & 41.9 & 70.73 & 57.90 & 80.06 & 66.75 & 80.41 & 71.24 & 88.45 & 79.29 \\
LViT~\cite{li2023lvit} & 29.70 & 54.1 & 74.52 & 61.15 & 76.64 & 62.12 & 83.40 & 74.78 & 90.63 & 82.86 \\
LGA~\cite{hu2024lga} & 8.24 & 381.1 & 74.53 & 61.10 & 80.55 & 67.43 & 83.46 & 74.50 & 89.85 & 81.56 \\
CMIRNet~\cite{xu2024cmirnet} & 239.80 & 134.5 & 73.88 & 60.25 & 79.64 & 66.17 & 79.63 & 69.87 & 87.45 & 77.71 \\
AT~\cite{zhong2023ariadne} & 44.00 & 22.4 & 72.41 & 58.63 & 78.53 & 64.65 & 79.57 & 69.48 & 87.19 & 77.30 \\
TeViA~\cite{11133474} & 146.80 & 11.2 & 72.32 & 59.06 & 79.47 & 65.94 & 84.12 & 75.69 & 90.96 & 83.41 \\
\midrule
\textbf{TIRNet (Ours)} & 41.10 & 24.1 & \textbf{75.41} & \textbf{62.77} & \textbf{81.38} & \textbf{68.61} & \textbf{84.77} & \textbf{76.47} & \textbf{91.23} & \textbf{83.88} \\
\bottomrule
\end{tabular}%
}
\end{table}
%
\section{Experiments}

\noindent\textbf{Datasets and Metrics.}
We evaluate TIRNet on two public datasets: MosMedData+~\cite{morozov2020mosmeddata} with 2,729 CT scans and QaTa-COV19~\cite{degerli2022osegnet} with 9,258 chest X-rays.
Following the data split protocol of LViT~\cite{li2023lvit}, we conduct extensive experiments on both datasets.
To quantitatively assess performance, we report both sample-level averages ($\mathrm{m\text{-}Dice}$, $\mathrm{m\text{-}IoU}$), and global metrics ($\mathrm{g\text{-}Dice}$, $\mathrm{g\text{-}IoU}$) aggregated over the entire dataset.
\begin{figure}[!t]
\centering
\includegraphics[width=1\textwidth]{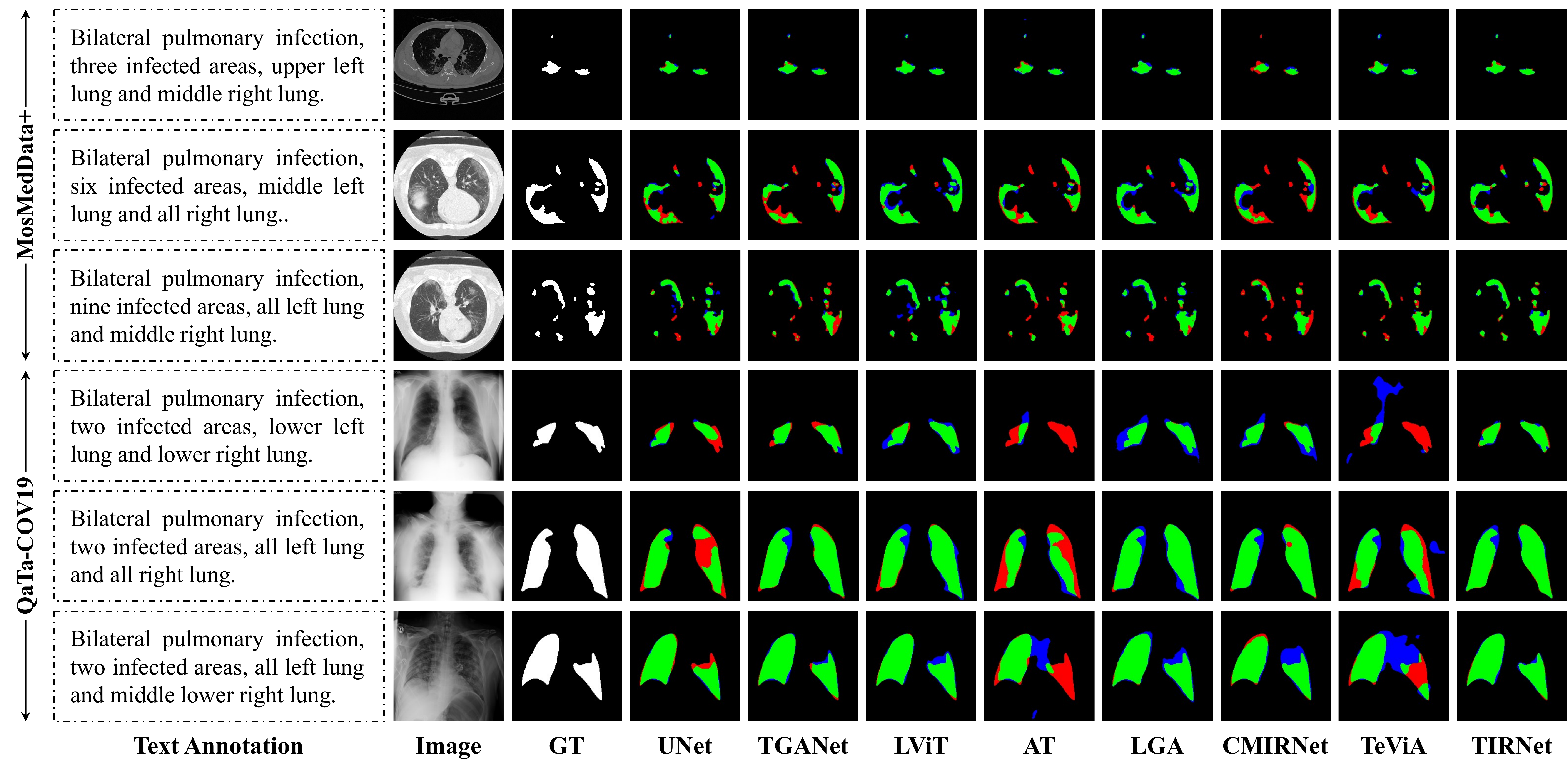}
\caption{Visual comparison of segmentation results of different methods (Green: True Positive, Red: False Negative, Blue: False Positive).}
\label{fig:comparison}
\end{figure}

\noindent\textbf{Implementation Details.}
TIRNet was implemented in PyTorch and trained on a single NVIDIA Tesla V100 GPU.
Input images were resized to $224 \times 224$ with a batch size of 32.
We used the AdamW~\cite{loshchilov2018decoupled} optimizer with an initial learning rate of $3 \times 10^{-4}$, decayed to $1 \times 10^{-6}$ via cosine annealing.
Data augmentation included
random flipping, rotation, scaling, Gaussian blur, and photometric distortions.

\noindent\textbf{Comparative Experiments.}
We compare TIRNet with existing methods on MosMedData+ and QaTa-COV19, with quantitative results reported in Table~\ref{tab:comparison}.
On MosMedData+, TIRNet improves over LViT~\cite{li2023lvit} by 0.89\% in $\mathrm{m\text{-}Dice}$ and 1.62\% in $\mathrm{m\text{-}IoU}$, and surpasses TeViA~\cite{11133474} by 1.91\% in $\mathrm{g\text{-}Dice}$ and 2.67\% in $\mathrm{g\text{-}IoU}$.
On QaTa-COV19, TIRNet also achieves the best overall performance, exceeding LViT~\cite{li2023lvit} by 1.37\% in $\mathrm{m\text{-}Dice}$ and 1.69\% in $\mathrm{m\text{-}IoU}$, and improving over TeViA~\cite{11133474} by 0.27\% in $\mathrm{g\text{-}Dice}$ and 0.47\% in $\mathrm{g\text{-}IoU}$.
The visual comparisons in Fig.~\ref{fig:comparison} further demonstrate that TIRNet produces more accurate and robust segmentation results across diverse challenging cases.
\begin{table}[tb]
\centering
\setlength{\tabcolsep}{1.5pt}
\renewcommand{\arraystretch}{1.05}
\begin{minipage}[t]{0.58\linewidth}
\centering
\scriptsize
\parbox[t][2.6\baselineskip][t]{\linewidth}{\caption{Ablation study results of key blocks and the MSIS-Loss.}\label{tab:AblationModule}}
\begin{tabularx}{\linewidth}{c c c *{4}{>{\centering\arraybackslash}X}}
\toprule
RTMB & CDCB & MSIS-Loss & $\mathrm{m\text{-}Dice}$ & $\mathrm{m\text{-}IoU}$ & $\mathrm{g\text{-}Dice}$ & $\mathrm{g\text{-}IoU}$ \\
\midrule
\ding{55} & \ding{55} & \ding{55} & 78.44 & 68.67 & 86.90 & 76.83 \\
\ding{51} & \ding{55} & \ding{55} & 83.13 & 74.45 & 90.48 & 82.61 \\
\ding{51} & \ding{55} & \ding{51} & 83.65 & 75.08 & 90.72 & 83.02 \\
\ding{55} & \ding{51} & \ding{55} & 79.97 & 70.53 & 87.71 & 78.11 \\
\ding{51} & \ding{51} & \ding{55} & 84.36 & 75.96 & 91.13 & 83.70 \\
\ding{51} & \ding{51} & \ding{51} & \textbf{84.77} & \textbf{76.47} & \textbf{91.23} & \textbf{83.88} \\
\bottomrule
\end{tabularx}
\end{minipage}\hfill
\begin{minipage}[t]{0.40\linewidth}
\centering
\scriptsize
\parbox[t][2.6\baselineskip][t]{\linewidth}{\caption{Sensitivity of hyperparameters $\lambda_{\mathrm{rgc}}$ and $\lambda_{\mathrm{bs}}$.}\label{tab:AblationWeight}}
\begin{tabularx}{\linewidth}{c *{4}{>{\centering\arraybackslash}X}}
\toprule
$\lambda_{\mathrm{rgc}} / \lambda_{\mathrm{bs}}$ & $\mathrm{m\text{-}Dice}$ & $\mathrm{m\text{-}IoU}$ & $\mathrm{g\text{-}Dice}$ & $\mathrm{g\text{-}IoU}$ \\
\midrule
$0 / 0$ & 84.36 & 75.96 & 91.13 & 83.70 \\
$1 / 0$ & 84.54 & 76.11 & 91.16 & 83.75 \\
$0 / 0.5$ & 84.20 & 75.81 & 91.22 & 83.85 \\
$0.5 / 0.5$ & 84.35 & 75.96 & 91.10 & 83.65 \\
$1 / 1$ & 84.39 & 75.94 & 91.04 & 83.55 \\
$1 / 0.5$ & \textbf{84.77} & \textbf{76.47} & \textbf{91.23} & \textbf{83.88} \\
\bottomrule
\end{tabularx}
\end{minipage}
\end{table}
\begin{figure}[tb]
\centering
\includegraphics[width=0.98\linewidth]{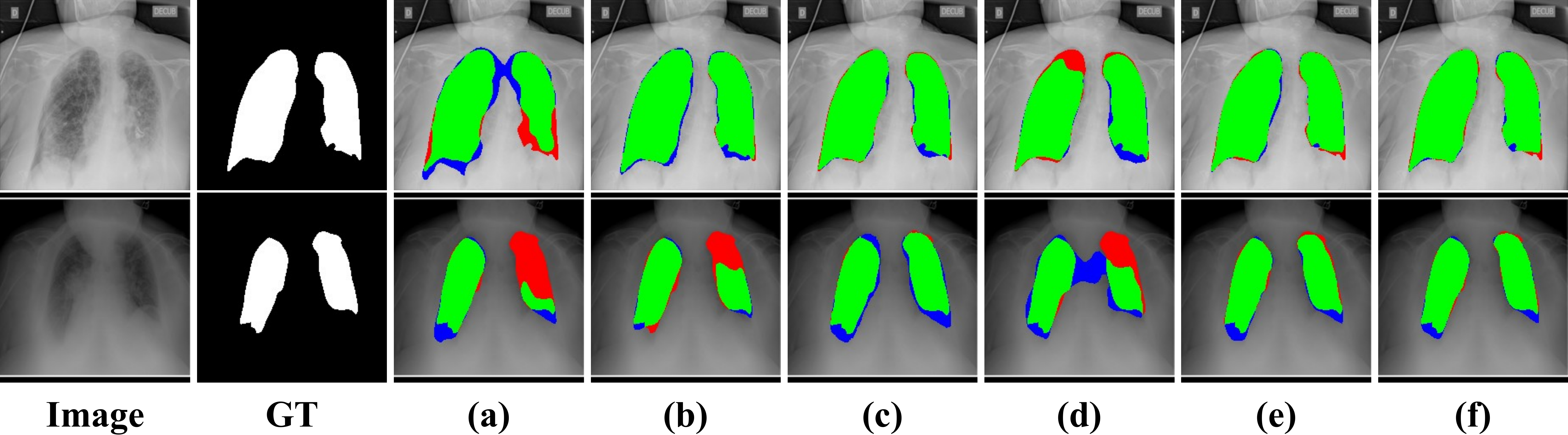}
\caption{Qualitative visualization of the component ablation study. Panels (a)--(f) correspond to the methods in the 1st--6th rows of Table~\ref{tab:AblationModule}, respectively.}
\label{fig:ablation}
\end{figure}

\noindent\textbf{Ablation Studies.}
To assess each component in TIRNet, we conducted ablation studies on the QaTa-COV19 dataset, with results summarized in Table~\ref{tab:AblationModule}.
Starting from the baseline, including the RTMB yields a 4.69\% increase in $\mathrm{m\text{-}Dice}$, showing that semantic illumination effectively enhances features.
The CDCB alone contributes a 1.53\% gain in $\mathrm{m\text{-}Dice}$, confirming its role in recovering high-frequency structural details.
Adding the MSIS-Loss to the RTMB further improves $\mathrm{m\text{-}Dice}$ by 0.52\%, highlighting the need for explicit cross-modal alignment.
Overall, the full TIRNet achieves the best performance, outperforming the baseline by 6.33\% in $\mathrm{m\text{-}Dice}$.
Fig.~\ref{fig:ablation} shows qualitative results for the component ablations in Table~\ref{tab:AblationModule}.
These results indicate that each component in TIRNet contributes to more accurate and semantically consistent segmentation.

In addition, we investigate the sensitivity of the hyperparameters $\lambda_{\mathrm{rgc}}$ and $\lambda_{\mathrm{bs}}$.
We evaluate different combinations of these hyperparameters, as reported in Table~\ref{tab:AblationWeight}.
The results indicate that the optimal performance is achieved when $\lambda_{\mathrm{rgc}}$ and $\lambda_{\mathrm{bs}}$ are set to 1 and 0.5, respectively.
\section{Conclusion}
In this paper, we propose TIRNet, which maps Retinex theory to LMIS.
Treating clinical text embeddings as semantic illumination, it separates foreground from similar backgrounds and reduces semantic mismatch in cross-modal interaction.
TIRNet consists of two key blocks, the RTMB and the CDCB.
The RTMB uses complementary positive and negative illumination maps to enhance features and suppress interference.
The CDCB restores high-frequency details via an illumination-guided consistency gate.
Furthermore, the MSIS-Loss aligns modalities by enlarging the foreground-background margin and matching negative illumination to the background distribution.
Experiments on the MosMedData+ and QaTa-COV19 datasets demonstrate that TIRNet achieves state-of-the-art performance, validating the effectiveness of mapping the Retinex illumination formulation to the cross-modal segmentation setting.
%
%
\bibliographystyle{splncs04}
\bibliography{Paper-1662}

\end{document}